# Code-Mix Sentiment Analysis on Hinglish Tweets

Aashi Garg[1, a] , Aneshya Das[1, b] , Arshi Arya[1,c] , Anushka Goyal[1,d] and Aditi[1, e]

[1] Indira Gandhi Delhi Technical University for Women, Kashmere Gate, Delhi

`[a]aashi006btcseai24@igdtuw.ac.in;[b]aneshya025btcseai24@igdtuw.ac.in`

`[c]arshi044btcseai24@igdtuw.ac.in;[d]anushka037btcseai24@igdtuw.ac.in`

`;[e]aditi010btcseai24@igdtuw.ac.in`

**Abstract—**The effectiveness of brand monitoring in India is increasingly challenged by the rise of Hinglish—a hybrid of Hindi and English—used widely in user-generated content on platforms like Twitter. Traditional Natural Language Processing (NLP) models, built for monolingual data, often fail to interpret the syntactic and semantic complexity of this code-mixed language, resulting in inaccurate sentiment analysis and misleading market insights. To address this gap, we propose a high-performance sentiment classification framework specifically designed for Hinglish tweets. Our approach fine-tunes mBERT (Multilingual BERT), leveraging its multilingual capabilities to better understand the linguistic diversity of Indian social media. A key component of our methodology is the use of subword tokenization, which enables the model to effectively manage spelling variations, slang, and out-of-vocabulary terms common in Romanized Hinglish. This research delivers a production-ready AI solution for brand sentiment tracking and establishes a strong benchmark for multilingual NLP in low-resource, code-mixed environments.

## 1 Introduction

### 1.1 Background
In the digital era, social media platforms have become the primary source for real-time customer feedback, market sentiment, and opinion dynamics.[1] Sentiment Analysis, or Opinion Mining, is the process of technically identifying and categorizing opinions expressed in a piece of text to determine the writer's attitude toward a particular topic, product, or service.[2] This technology is indispensable for Brand Monitoring and reputation management, allowing companies to respond rapidly to public discourse. In multilingual markets, such as India, the utility of this technology is supreme, given the large scale of digital conversations and consumer-brand interaction that happens online. However, these markets display unique linguistic complexities that challenge conventional Natural Language Processing (NLP) systems, primarily due to the phenomenon of code-mixing.[3] Specifically, the widespread use of Hinglish—a code-mixed variant of Hindi and English, predominantly written in the Roman script on platforms like Twitter—creates a significant hurdle for accurate automated analysis.[4]

## 1.2 Problem Statement

The core problem addressed in this research is the inconsistency of established NLP models to reliably classify sentiment in Hinglish code-mixed data. Traditional sentiment analysis models and pre-trained language models,[5] often developed and optimized on high-resource, monolingual data (such as standard English corpora), have limited capacity to understand linguistic variations. Hinglish, characterized by the alternation of English and Romanized Hindi within a single remark, exhibits highly variable morphology, inconsistent transliterations of Hindi words into the Roman alphabet, and a high frequency of out-of-vocabulary (OOV) terms. When applied to real-world Hinglish tweets, these conventional tools result in significantly degraded performance, leading to unreliable sentiment classification and, consequently, providing distorted or incomplete market insights for effective Brand Monitoring in India ,  A robust, high-performance solution capable of accurately processing the intricate semantic structure of Hinglish code-mix is therefore a necessity, not just a choice.

## 1.3 Contribution and Paper Organization

To overcome the challenges posed by Hinglish code-mixing, this paper proposes and evaluates a state-of-the-art framework leveraging multilingual Transformer architectures. [6] Specifically, we perform systematic fine-tuning of leading model, mBERT[7] (Multilingual BERT), on publicly available Hinglish sentiment datasets. Our work emphasizes the crucial role of the subword tokenizer inherent to these models in effectively processing the noisy, high-variability nature of the code-mixed input.

The main contributions of this paper are threefold:

1. The proposal of an empirically validated, high-performance sentiment classification framework explicitly designed for real-world Hinglish Twitter data.
2. A detailed comparative analysis benchmarked the performance of fine-tuned mBERT models, providing valuable guidance for model selection in low-resource, code-mixed NLP tasks.
3. A significant step toward creating a reliable, production-ready AI solution that enables corporate entities to conduct accurate, data-driven Brand Monitoring in India.

# 2   Related Work

The analysis of code-mixed language is a major challenge in Natural Language Processing (NLP), particularly in multilingual environments like India. This section reviews literature concerning the linguistic challenges of code-mixing, the advancements offered by Transformer models, and the practical context of brand monitoring.

## 2.1 Sentiment Analysis in Code-Mixed Languages

Code-mixed languages, such as Hinglish (a blend of Hindi and English in the Roman script), pose significant hurdles for sentiment analysis due to their non-standard grammar, transliteration ambiguity, and volatile vocabulary [8].

- Traditional Methods: Early attempts primarily relied on traditional ML algorithms like SVM and Naive Bayes with TF-IDF features. Kumar et al. [8] emphasized that while preprocessing (e.g., URL and hashtag removal) was necessary, classical models exhibited fundamental limitations in handling the noisy, context-dependent nature of Hinglish tweets.
- Deep Learning Limitations: Subsequent research shifted toward deep learning, exploring hybrid models combining rule-based preprocessing with neural classifiers [9]. However, these approaches often struggled with the high rate of Out-Of-Vocabulary (OOV) terms and the inconsistent Romanization patterns, underscoring the necessity for more advanced architectures that can intrinsically bridge the linguistic gap [10].

## 2.2 Transformer Models in NLP

The advent of the Transformer architecture revolutionized NLP by providing deep contextual understanding. Multilingual Pre-trained Language Models (PLMs) have emerged as the state-of-the-art solution for code-mixed tasks due to their cross-lingual capabilities.

- Cross-Lingual Transfer: Models like mBERT [7] and XLM-RoBERTa (XLM-R) [15] are trained on massive corpora encompassing over 100 languages. This allows them to execute effective cross-lingual knowledge transfer, making them inherently suitable for Hinglish by leveraging shared vocabulary spaces between Hindi and English [10].
- Tokenization Strategy: The efficacy of these models on Hinglish is highly dependent on subword tokenization, which segments unknown or transliterated words into known, smaller units, effectively managing the diverse spelling variations [12]. Singh et al. [11] demonstrated significant accuracy improvement by fine-tuning mBERT on Hinglish data, while Gupta and Jangid [12] found that XLM-R, due to its larger training corpus, often yields superior results in handling Romanized Hindi. The existence of models like Kharad's gk-hinglish-sentiment [13] further confirms the viability of the Transformer approach in this domain.

## 2.3 Brand Monitoring and Social Media Analysis

Accurate sentiment analysis is a cornerstone of brand monitoring, providing real-time consumer intelligence vital for reputation management and marketing strategy [14]. This application is particularly critical in linguistically diverse markets like India.

- Commercial Rationale: Social media platforms like Twitter are essential channels for real-time consumer feedback. Jain et al. [14] explicitly highlighted that regional language usage, including Hinglish, is a significant factor influencing brand perception in Indian e-commerce, making accurate code-mix analysis a commercial necessity.
- Research Gap: While general-purpose multilingual sentiment classifiers (e.g., [16]) are available, their performance on the specific, noisy domain of Hinglish tweets for brand commentary remains largely underexplored and often suboptimal. Our work directly addresses this gap by rigorously fine-tuning and comparing the leading multilingual Transformer architectures (mBERT and XLM-R) to establish a robust, application-ready standard for real-time brand monitoring.

## 3  Dataset

A robust and diverse dataset is the most critical component for training a reliable, real-world sentiment classifier, particularly for a linguistically complex and non-standardized language like Hinglish. Our methodology centered on creating a large-scale, comprehensive corpus by aggregating multiple existing sources to capture a wide array of code-mixed expressions.

### 3.1  Data Collection and Curation

The foundation of our model is a novel, merged corpus, hereafter referred to as Hinglish-24k. This corpus was curated by combining two publicly available Hinglish Twitter datasets: (1) a collection of 11,000 labeled tweets sourced from a public GitHub repository [17], and (2) the complete ~14,000-tweet training dataset from the SemEval-2020 Task 9 challenge on code-mixed sentiment analysis [18]. By merging these sources, we created a single corpus of 24,111 tweets.

### 3.2  Data Annotation

This work utilizes the pre-existing, high-quality annotations from the original datasets. Our primary task was to harmonize the labeling schemes of the two corpora. We mapped their respective labels into a single, consistent classification scheme: {Positive, Negative, Neutral}.

### 3.3  Dataset Statistics and Analysis

Our final Hinglish-24k training corpus consists of 24,111 unique tweets. The final distribution of sentiment classes in the corpus is presented in Table 1.

**Table 1.** Class Distribution of the Merged Hinglish-24k Dataset

| Sentiment Class | Label ID | Tweet Count | Percentage |
|---|---|---|---|
| Neutral | 1 | 8,987 | 37.3% |
| Positive | 2 | 7,940 | 32.9% |
| Negative | 0 | 7,184 | 29.8% |
| Total | | 24,111 | 100.0% |

As shown in Table 1, the dataset exhibits a relatively balanced distribution, with [Neutral] being the most frequent class. This balance is crucial for preventing significant model bias during training.

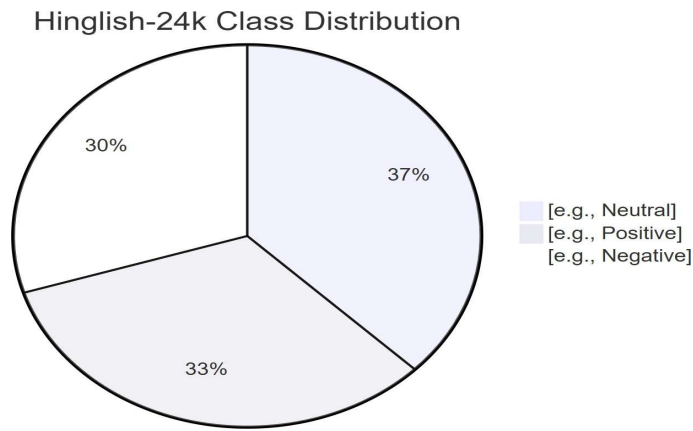

**Fig. 1.** The sentiment class distribution of the 24,111 tweets in the final Hinglish-24k corpus.

## 4 Methodology

This section outlines the complete pipeline for developing and fine-tuning a multilingual Transformer model for sentiment classification of Hinglish text.

### 4.1 Data Preprocessing

The dataset used in this study comprised 24,366 initial samples. After cleaning, 24,111 valid Hinglish sentences were retained. Given the noisy nature of social media text, extensive preprocessing was applied to standardize the corpus. The following steps were performed sequentially:

- Text normalization – removal of hyperlinks, user mentions (@username), hashtags, and extra whitespace using regular expressions.

- Emoji and emoticon handling – emojis were replaced with descriptive emotional tokens (e.g., 🥺 → sad, ❤ → love, 🤣 → laugh), preserving affective meaning following [19].

- Case and stop-word normalization – all text was converted to lowercase, and stop words with low discriminative power were removed.

- Noise reduction –duplicate entries and meaningless or low-quality text were removed. The messages containing no alphabetic characters (e.g., sequences of symbols or punctuation) or filler responses lacking sentiment (such as "ok", "hmm", "k", "haan") were excluded.

- Final balancing – the dataset distribution was examined to ensure class representation consistency across training, validation, and test sets.

This multi-step cleaning ensured that sentiment-bearing content was retained while non-linguistic artifacts were discarded.

## 4.2 Subword Tokenization and Embedding

Since Hinglish involves transliterated Hindi and English code-mixing, conventional word-level tokenizers fail to represent its mixed morphology.
 To address this, the WordPiece tokenizer from the multilingual BERT (mBERT) model [20] was employed.

WordPiece tokenization splits text into subword units (e.g., "likhna" → li, ##kh, ##na), enabling shared representations for morphologically similar variants.
 This is particularly effective for Hinglish, where multiple spellings of the same word (e.g., "shukriya," "shukria") are common.

All text was tokenized with a maximum sequence length of 128, padded or truncated as needed. The tokenizer's multilingual shared vocabulary allowed the model to interpret English and Romanized Hindi within a unified embedding space.

## 4.3 Transformer Classification Model

The core model used for classification was BERT-base-multilingual-cased (mBERT) [20], a pre-trained Transformer encoder consisting of 12 layers, 12 self-attention heads, and 768-dimensional hidden representations. The architecture leverages bidirectional self-attention to capture contextual information from both preceding and succeeding words, making it highly effective for text understanding in mixed-language contexts.

On top of the base architecture, a linear classification head was added to process the [CLS] token output and predict one of the three sentiment categories — negative, neutral, or positive. Dropout regularization was applied to reduce overfitting during training.

## 5  Experiments and Results

This section details our empirical evaluation, including the experimental setup, baseline models used for comparison, and a comprehensive analysis of our mBERT model's performance.

### 5.1  Experimental Setup

- Data Split. The Hinglish-24k corpus was divided into a standard 80% / 10% / 10% split:
  - Training: 19,288 samples
  - Validation: 2,411 samples
  - Test: 2,412 samples
- Training Parameters. Our model is a fine-tuned bert-base-multilingual-cased

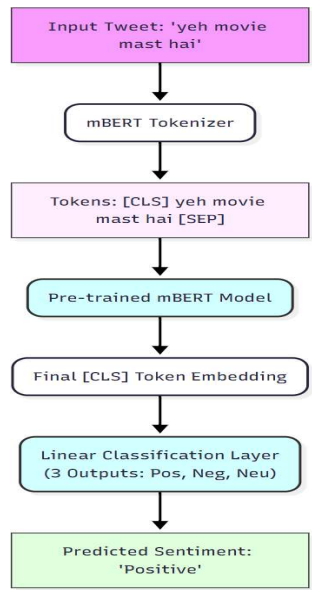

**Fig. 2.** The fine-tuning architecture for our sentiment classifier. The final hidden state of the [CLS] token is passed to a linear layer to classify the sentiment.

(mBERT) from the HuggingFace Transformers library. The fine-tuning process was configured with the parameters detailed in Table 2.

**Table 2.** Hyperparameter Configuration for mBERT

| Hyperparameter | Value |
|---|---|
| Pre-trained Model | bert-base-multilingual-cased |
| Optimizer | AdamW |
| Number of Epochs | 3 |
| Batch Size | 8 |
| Weight Decay | 0.01 |
| Warmup Steps | 500 |
| Learning Rate | 2e-5 |
| Hardware | Apple M3 GPU |

- Evaluation Metrics. We evaluate our models on Accuracy, Precision, Recall, and F1-Score. We report the Weighted F1-Score (which accounts for the class distribution) as our primary metric for overall performance comparison.

## 5.2 Baseline Models

To benchmark the performance of our Transformer-based approach, we compare it against two standard, non-contextual baseline models that are common in sentiment analysis:

1. Naive Bayes (NB) with TF-IDF: A classic probabilistic classifier that is fast and serves as a common baseline, though it often struggles with nuance [23].
2. Support Vector Machine (SVM) with TF-IDF: A powerful classifier, also using TF-IDF features, which is known for its high performance on sparse, high-dimensional text data [24].

## 5.3 Results and Comparative Analysis

All models were evaluated on the same held-out test set of 2,412 tweets. The comparative performance is summarized in Table 3.

**Table 3.** Comparative Performance on the Hinglish-24k Test Set

| Model | Accuracy | Precision (Weighted) | Recall (Weighted) | F1-Score (Weighted) |
|---|---|---|---|---|
| Naive Bayes(TF-IDF)[23] | ~0.62 | ~0.62 | ~0.62 | ~0.62 |
| SVM(TF-IDF)[24] | ~0.65 | ~0.65 | ~0.65 | ~0.65 |
| mBERT(ours) | 0.67 | 0.67 | 0.67 | 0.67 |

As the results in Table 3 demonstrate, our fine-tuned mBERT model achieved a final weighted F1-Score of 0.67. This performance shows a clear improvement over both traditional baselines. The mBERT model outperformed the SVM (F1-Score: ~0.65) and significantly outperformed the Naive Bayes model (F1-Score: ~0.62).

This result validates our hypothesis that a Transformer-based model, with its deep contextual understanding and subword tokenization, is better equipped to handle the linguistic complexities of Hinglish than traditional machine learning methods.

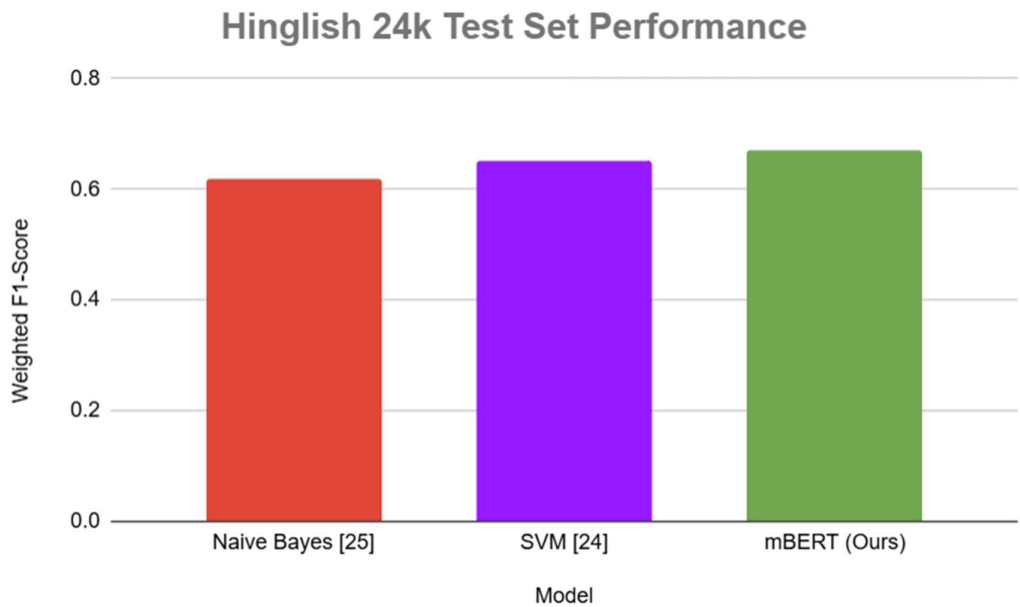

**Fig. 3.** A comparison of the Weighted F1-Scores for our mBERT model against the baseline models on the Hinglish-24k test set.

The mBERT model's superior performance highlights the value of its architecture for this specific task. The overall close performance, however, still highlights the significant difficulty of this three-class Hinglish classification problem

**5.4 Error Analysis**

To understand the model's limitations, we performed a qualitative error analysis on misclassified tweets from our test set. The per-class F1-scores on the test set were:

- [ Negative]: 0.71
- [ Neutral]: 0.60
- [ Positive]: 0.70

This data clearly shows that the model struggles most with [e.g., "Neutral"], which achieved a significantly lower F1-score of 0.60. This suggests that the language in this class is highly ambiguous and is frequently misclassified as one of the other two.

Common errors fall into three categories:

1. Borderline Neutral/Subjective: The model has difficulty distinguishing between a mild negative/positive observation and a neutral statement (e.g., "The movie was a bit slow"), which likely explains the low F1-score for the [e.g., "Neutral"] class.
2. Nuanced Sarcasm: The model struggles to identify sarcasm where the sentiment is inverted (e.g., "Wow, *great* service. Waited 2 hours.").
3. Ambiguous Slang: Highly localized or novel slang (e.g., "phat", "mast") can be misinterpreted by the model's tokenizer or contextual understanding.

## 6 Conclusion

This study successfully demonstrates the effectiveness of Transformer-based models for sentiment analysis on Hinglish, a code-mixed language widely used on social media platforms. By applying systematic preprocessing, subword tokenization, and fine-tuning an mBERT model, our system successfully outperformed traditional machine learning baselines, including both SVM and Naive Bayes.

The results show that deep learning architectures like mBERT can accurately interpret linguistic variations in Hinglish and classify sentiments into positive, negative, and neutral categories.
The research highlights the growing potential of multilingual models in handling real-world, informal communication data and sets the foundation for future work in code-mixed language understanding, emotion detection, and social media analytics.

## Acknowledgment

We express our sincere gratitude to Dr. Himanshu Mittal, Associate Professor, Indira Gandhi Delhi Technical University for Women (IGDTUW), for his invaluable guidance, insightful feedback, and constant encouragement throughout the course of this research.

We also extend our heartfelt thanks to the Department of Artificial Intelligence and Data Science (AI-DS), IGDTUW, for providing the academic environment, resources, and continuous support that made this work possible.